\documentclass[11pt]{article}

\usepackage[final]{acl}

\usepackage{times}
\usepackage{latexsym}

\usepackage[T1]{fontenc}
\usepackage[utf8]{inputenc}

\usepackage{microtype}

\usepackage{inconsolata}

\usepackage{graphicx}

\usepackage{url}
\usepackage{caption}
\usepackage{subfig}
\usepackage{multirow}
\usepackage{makecell}
\usepackage[inline]{enumitem}
\usepackage{hyperref}
\usepackage{amsmath}
\usepackage{listings}
\usepackage{bold-extra}
\usepackage{subcaption}
\usepackage{subfig}
\usepackage{float}

\title{Time Is Effort: Estimating Human Post-Editing Time for Grammar Error Correction Tool Evaluation}

\author{
  Ankit Vadehra$^{\ast}$ 
  \\
  University of Waterloo, Vector Institute
  \\
  \texttt{avadehra@uwaterloo.ca}
  \And
  Bill Johnson 
  \\
  Scribendi Inc.
  \\
\texttt{bill.johnson@scribendi.com}
  \AND
  Gene Saunders 
  \\
  Scribendi Inc.
  \\
\texttt{gene.saunders@scribendi.com}
  \And
  Pascal Poupart 
  \\
  University of Waterloo, Vector Institute
  \\
  \texttt{ppoupart@uwaterloo.ca}
}
\begin{document}
\maketitle
\renewcommand{\thefootnote}{\fnsymbol{footnote}}
\footnotetext[1]{Corresponding author.}
\renewcommand{\thefootnote}{\arabic{footnote}}
\begin{abstract}
Text editing can involve several iterations of revision. Incorporating an efficient Grammar Error Correction (GEC) tool in the initial correction round can significantly impact further human editing effort and final text quality. This raises an interesting question to quantify GEC Tool usability: How much effort can the GEC Tool save users? We present the first large-scale dataset of post-editing (PE) time annotations and corrections for two English GEC test datasets (BEA19 and CoNLL14). We introduce Post-Editing Effort in Time (PEET) for GEC Tools as a human-focused evaluation scorer to rank any GEC Tool by estimating PE time-to-correct. Using our dataset, we quantify the amount of time saved by GEC Tools in text editing. Analyzing the edit type indicated that determining whether a sentence needs correction and edits like paraphrasing and punctuation changes had the greatest impact on PE time. Finally, comparison with human rankings shows that PEET correlates well with technical effort judgment, providing a new human-centric direction for evaluating GEC tool usability.\footnote{We release our dataset and code at - \url{https://github.com/ankitvad/PEET_Scorer}}
\end{abstract}

\section{Introduction}
Grammar Error Correction (GEC) is an important step of the text editing process. There has been a lot of work to build automated GEC tools that can improve the structure and fluency of text while also correcting language errors \cite{bryant2023grammatical}. Since GEC tool-assisted text editing is an iterative process, an editor can make post-edits to the tool output to obtain the closest targeted correction. Estimating the post-editing (PE) effort required to reach the targeted correction can be used as a quality evaluation for the tool.

Human-in-the-loop PE effort was introduced and explored extensively for Machine Translation (MT) \cite{koponen2016machine} systems. PE effort is studied across three levels \cite{krings2001repairing}: technical effort, which is the number of edits; cognitive effort, which denotes the psychological assessment required to identify and correct the errors; and temporal effort, which is the total time taken to evaluate and perform post-edits (which includes technical and cognitive effort). \citet{ye2021predicting} and \citet{tezcan2019estimating} have explored estimating MT PE time based on edit features. Technical PE effort has also been studied in areas like Text Summarization \cite{lai-etal-2022-exploration}, Natural Language Generation \cite{sripada2004evaluating} and GEC \cite{rozovskaya2021good, ostling2023evaluation}.

To incorporate the human editor effort in text correction, we present the first work to consider PE effort in Time (PEET) scores for quality estimation of a GEC tool. The usability of a GEC tool depends inversely on the PE effort to fix the tool output. We release the first large-scale dataset capturing time-to-correct annotations for two English GEC test sets - BEA19 \cite{bryant2019bea} and CoNLL14 \cite{ng2014conll}, post-edited from two conditions: the original sentence and the output from two strong GEC tools - GECToR \cite{omelianchuk2020gector} and GEC-PD \cite{kiyono2019empirical}. We further present a new human-centric GEC Tool evaluation method - PEET Scorer, to estimate the time-to-correct for GEC Tool predictions, which correlates well with human editing effort. As a result, we propose that the PEET scorer can be incorporated along with Post-Editing to evaluate a GEC Tool from a human editor's perspective.

In this work, we make the following contributions:
\begin{enumerate}
    \item We present the first large-scale GEC dataset with post-editing time-to-correct annotations along with three new high-quality human-preference targeted correction sets for two GEC Test datasets (BEA19 and CONLL14) - source sentence correction and post-edit for two strong GEC Tools (GECToR and GEC-PD) output.
    \item We quantify the editing time saved and improvement in final correction quality (estimated using GEC metrics) using GEC Tools for first-pass text-editing. We also observe that determining whether a sentence needs correction and edits like paraphrasing and punctuation changes has the greatest impact on time-to-correct.
    \item We contribute a new evaluation method called PEET Scorer that can be used to rank any GEC Tool in terms of time-to-correct.  We compare the PEET scorer with 3 human judgment rankings of 33 GEC Tools, and demonstrate high correlation with further correction effort required.
\end{enumerate}

\section{Background Work}
\subsection{Grammar Error Correction (GEC) Tools}\label{sec:gec}
GEC tools can be broadly divided into supervised-trained, LLM-based, and ensemble-ranked models \cite{omelianchuk2024pillars}.

The supervised GEC tools can be divided into edit-based and sequence-to-sequence models. % similar to the neural machine translation (NMT) architecture \cite{vaswani2017attention}.
Edit-based models convert the task to a sequence-tagging and editing approach where each token in the input sentence is assigned an edit operation. Some tools that use this approach are the PIE \cite{awasthi2019parallel} and GECToR \cite{omelianchuk2020gector, tarnavskyi2022ensembling} models. Sequence-to-Sequence (S2S) GEC Tools utilize an encoder-decoder architecture where the corrected sentence is generated for each input sentence \cite{choe2019neural, grundkiewicz2019neural, kiyono2019empirical}.

Large language models like Llama \cite{touvron2023llama, omelianchuk2024pillars} and ChatGPT \cite{katinskaia2024gpt} also perform well for GEC \cite{zhang2023multi, fang2023chatgpt} in different settings like - Zero-Shot, Few-Shot and Fine-Tuning \cite{korniienko2024enhancing, davis2024prompting, raheja2023coedit}. The current state-of-the-art GEC tools all rely on the approach of ensembling multiple strong GEC Tools, aggregating them with methods like majority votes \cite{tarnavskyi2022ensembling} and logistic regression \cite{qorib2023system,qorib2022frustratingly}.

In this work, we use two supervised GEC tools for first-pass text editing: GECToR edit tagging \cite{omelianchuk2020gector} and GEC-PseudoData (GEC-PD) \cite{kiyono2019empirical} model, which was trained on a large synthetic corpus. The output of these models is further corrected by human editors while tracking the time-to-correct (temporal effort). We use this time dataset to quantify the impact of GEC tools for text-editing, observing reduced post-editing time and better quality final correction (Section \ref{sec:impactgec}). Even though the GEC Tools we selected (GECToR and GEC-PD) are not the most recent, they are on par with human-level performance as demonstrated in Section \ref{sec:correct-qual} - Table \ref{tab:errant_scores_all_ourTRG}.

\subsection{Post Editing Effort in Machine Translation}\label{sec:PEE-MT}
Post Editing Effort (PEE) for Quality Estimation is an actively researched task in Machine Translation (MT). It evaluates the output of an MT system for quality and correctness \cite{senez1998post, specia2011exploiting}. Post-editing (PE) the output of an MT system can improve the final translation quality compared to translating the source from scratch, while improving overall editor productivity \cite{plitt2010productivity, guerberof2009productivity, green2013efficacy}. We briefly review previous work in MT that explores PEE across three levels (technical, cognitive and temporal effort) \cite{krings2001repairing}.

Technical effort has been defined by edit distance metrics like - Translation Edit Rate (TER) and Human TER \cite{snover2006study} as well as keystroke and edit operation logging \cite{barrachina2009statistical, o2005methodologies, carl2011process}. Cognitive effort has also been studied in terms of edit complexities \cite{temnikova2010cognitive, koponen2012post, popovic2014relations, daems2017identifying} and human-assessed quality judgment and ranking \cite{specia2009estimating, specia2011predicting, koponen2012comparing}. Keystroke logs to determine pause information \cite{o2005methodologies, carl2011process}, eye gaze tracking and pause fixation \cite{vieira2014indices, hvelplund2014eye, daems2015impact} and Thinking Aloud Protocol (TAP) \cite{krings2001repairing, vieira2017cognitive, o2005methodologies} have also been proposed as measures of cognitive effort. The work on Temporal Effort in MT estimates the relationship between the time-to-correct and different evaluation metrics \cite{tatsumi2009correlation}, source/target translation characteristics \cite{tatsumi2010source}, and quality estimation \cite{specia2011exploiting}.
%To the best of our knowledge (at the time of writing this paper), there has been limited work that estimates PE temporal effort.
\citet{zaretskaya2016measuring} and \citet{popovic2014relations} study the average temporal effort required for each error type by considering the time-to-correct and frequency of error edits. Finally, \citet{ye2021predicting} and \citet{tezcan2019estimating} train models to estimate the post-editing time based on PE features. 

PE has also been explored in tasks like Text Summarization \cite{lai-etal-2022-exploration} and  Cognitive and Technical PE Effort has been studied for Grammar Error Correction (GEC) evaluation.

\subsection{Post Editing Effort in Grammar Error Correction}\label{sec:PEE-GEC}
We review previous work in GEC that closely relates to post-editing (PE) effort across two levels (cognitive and technical effort). To the best of our knowledge, temporal effort for PE has not been explored for GEC tools.
\subsubsection{Cognitive Post Editing Effort}
Although cognitive PE effort has not been measured directly for GEC, Human judgment rankings of GEC Tools \cite{grundkiewicz2015human, kobayashi2024revisiting, napoles2019enabling}, which are an estimate of perceived cognitive effort, have been used extensively for GEC evaluation metric assessment. Reference-based GEC metrics like ERRANT \cite{bryant2017automatic}, $M^{2}$ \cite{dahlmeier2012better}, GoToScorer \cite{gotou2020taking}, and GLEU \cite{courtney2016gleu} and reference-less metrics like PT-$M^2$ \cite{gong2022revisiting}, Scribendi Score \cite{islam2021end}, SOME \cite{yoshimura2020some} and IMPARA \cite{maeda2022impara} designed to estimate GEC Tool quality are trained and evaluated using the GEC human judgment rankings.

However, perceived cognitive effort does not always agree with the actual PE effort and can be subjective. Sentence correction experiments in GEC have shown poor cognitive agreement between editors. \citet{tetreault2014bucking} and \citet{tetreault2008native} asked 2 native English speakers to insert a preposition into 200 sentences, from which a single preposition was removed, obtaining an agreement score of just $0.7$. \citet{rozovskaya2010annotating} asked three annotators to evaluate and mark 200 sentences for correctness, showing a poor pairwise agreement between them ($0.4, 0.23, 0.16$). Finally, there has been some work considering the cognitive proficiency of the user interacting with a GEC Tool \cite{nadejde2020personalizing} and the annotators who create the evaluation references of GEC test sets \cite{takahashi2022proqe, napoles2017jfleg}.

Surprisingly, none of the GEC metrics described above have considered using targeted references (target obtained after correcting the GEC Tool output) to estimate the tool usability dependent on human PE effort.
\subsubsection{Technical Post Editing Effort}
To the best of our knowledge, only two prior studies have explored the impact of PE technical effort on GEC evaluation. \citet{rozovskaya2021good} introduced targeted references for English and Russian datasets and \citet{ostling2023evaluation} utilize PE references to assess Swedish GEC Tools. The studies show that GEC evaluation using untargeted references ignores the human subjectivity involved in text correction. For instance, the SEEDA - human judgment rankings from \citet{kobayashi2024revisiting} compared the correction outputs of GPT3.5, human editors and various Neural GEC Tools. The GPT-3.5 and human corrections were ranked significantly higher and contained nearly two and three times more edits than other corrections. As a result, these high-quality corrections obtain poor evaluation scores when compared against untargeted references. This inconsistency highlights the importance of PE for GEC Tool evaluation, to capture the true technical effort. 

Apart from estimating the PE effort, targeted references can also be used for fine-tuning and aligning Large Language Models (LLMs) with human preferences to generate better outputs \cite{li2024dissecting}.

\subsubsection{Temporal Post Editing Effort}
We introduce the first work to study the Temporal Effort in PE for GEC. Temporal effort described in terms of time-to-correct can efficiently capture the overall PE effort. We present the first large-scale dataset of post-edited corrections along with their temporal effort annotations for two strong GEC tools, GECToR \cite{omelianchuk2020gector} and GEC-PD \cite{kiyono2019empirical}, outputs on two English GEC Test sets - CONLL14 \cite{ng2014conll} and BEA19 \cite{bryant2019bea}. We also use this dataset to quantify the impact of GEC Tools in Text Editing and the contribution of different edit types to the human post-editing effort. We present PEET Scorer, a regression-based metric, to estimate the time-to-correct scores, which can be incorporated along with post-editing to evaluate the usability of GEC Tools in a human-centred manner.

\section{Dataset Collection and Processing}
\label{sec:dataInfo}
An important component in this work is the high-quality dataset of post-edit corrections for GEC, along with their time-to-correct (temporal effort) annotations. We partnered with a professional text-editing company - Scribendi Inc.\footnote{\url{https://www.scribendi.com/}} to collect this data. This section explains our dataset collection, filtering, and quality estimation process.

\subsection{Dataset Source}
We use source sentences from two popular English GEC test sets - CONLL14 \cite{ng2014conll} and BEA19 \cite{bryant2019bea} ($1312+4477=5789$ sentences). Each sentence was corrected in three variations: the source and post-editing outputs from Two GEC Tools - GECToR \cite{omelianchuk2020gector} and GEC-PD \cite{kiyono2019empirical} (Section \ref{sec:gec}). Each sentence variation was corrected by 1 out of 8 professional text editors, employed by Scribendi Inc. This resulted in a dataset of $5789*3=17367$ target corrections along with their time-to-correct scores.

\subsection{Editor Correction Framework}\label{sec:framework}
The source sentence and GEC Tool output serve as the basis for further editor correction. This follows the real framework for Text Editing, where a GEC Tool output is evaluated for further correction, compared with the original sentence. The editors were given GEC post-editing (PE) instructions (Appendix \ref{app:survey}-\ref{fig:surv_inst}) and asked to perform minimal edits and avoid rewrites. We used the Qualtrics\footnote{ \url{https://www.qualtrics.com/}} survey tool to collect PE corrections and enabled the "Timing Question" to log time-to-correct for each source sentence. All other metadata logging was disabled.

The 3 variations for each sentence - source, GECToR and GEC-PD output- were given to a different professional editor (in a pool of 8 editors) to eliminate any time-to-correct bias. The task of evaluating $17,367$ sentences was performed in batches of $50$. The editors were shown the source sentence and the first-pass GEC Tool output (Appendix \ref{app:survey}-\ref{fig:survey_example}). The final target correction and time-to-correct were logged for each sentence. For source sentence correction, only the original sentence was presented.

\subsection{Data Filtering}

To improve the dataset quality, we perform two stages of data filtering on the 3 target correction sets for each source ($17367$ sentences initially). In the first stage, we eliminate outliers based on the logged time-to-correct. \citet{snover2006study} showed that editors took between 3-6 minutes for each correction. Considering this and the distribution of the time-to-correct in our dataset, we filter corrections that took more than 250 seconds ($17033$ sentences remaining). Finally, we merge duplicate corrections from our dataset by averaging the time-to-correct values ($14112$ sentences dataset). This filtering allows us to retain $81.26\%$ of our dataset that we use as train and test sets (80:20 split) for the Post-Editing Effort in Time (PEET) Scorer.

\subsection{Correction Quality}\label{sec:correct-qual}
We collect and present three new target corrections for the CONLL14 \cite{ng2014conll} and BEA19 \cite{bryant2019bea} test datasets. The correction for the source and two post-edited target corrections. We evaluate the quality of the target corrections using the official GEC competition metric and the Inter Annotator Agreement (IAA) scores. Each target correction set can be divided into CONLL14 and BEA19 corrections. We evaluate the CONLL14 and BEA19 target corrections separately.

\begin{table}[h!]
\centering
\begin{tabular}{ |c c c|}
\hline
Correction & M2 Score & (Precision : Recall)\\
\hline
A1 & 46.9 & 44.6 : 59.1\\
A2 & 53.0 & 51.7 : 59.5\\
A3* & 98.6 & 98.7 : 98.3\\
A4 & 55.3 & 54.9 : 57.0\\
A5 & 52.8 & 51.3 : 59.7\\
A6 & 56.4 & 55.8 : 58.8\\
A7* & 98.6 & 98.7 : 98.5\\
A8 & 53.5 & 53.8 : 52.6\\
A9 & 55.7 & 55.6 : 56.0\\
A10 & 52.8 & 51.3 : 59.4\\\hline
c1 & 50.9 & 49.0 : 60.4\\
c2 & 52.3 & 50.5 : 61.0\\
c3 & 53.7 & 52.1 : 60.8\\
\hline
\end{tabular}
\caption{The M2 precision and recall quality score for all \citet{bryant2015far} target correction sets for the official CONLL14 competition task.}
\label{tab:c14-m2res}
\end{table}

\citet{bryant2015far} released 10 additional target corrections for the CONLL14 test dataset. We compare the quality scores of our 3 corrections with theirs using the official CONLL14 competition - M2 Scorer \cite{ng2014conll} metric. Table \ref{tab:c14-m2res} shows the M2 scores for all target correction sets - \citet{bryant2015far} corrections $A1-A10$, and our corrections $c1-c3$. Corrections A3 and A7 obtain near-perfect quality scores, since they were generated by the 2 editors who created the official CONLL14 competition target references \cite{bryant2015far}.
Ignoring the 2 outliers, we observe similar quality scores for our corrections. This indicates that our 3 CONLL14 Target corrections are of similar high quality. Unfortunately, there are no public correction references available for the BEA19 Test set (this work being the first to present 3 target references), making it hard to compare the quality scores directly.

\begin{table*}[t]
\centering
\begin{tabular}{|l|c|c|}
\hline
\textbf{\makecell{Candidate Set}} & \makecell{\textbf{BEA19 Test}\\(P : R : $F_{0.5}$)} & \makecell{\textbf{CONLL14 Test}\\(P : R : $F_{0.5}$)} \\ \hline\hline
\textbf{\makecell{Source Sentence}} & - & -\\
\textbf{\makecell{Source Sentence EC}} & 45.30 : 66.08 : 48.34 & 49.05 : 60.45 : 50.97\\ \hline
\textbf{\makecell{GECToR Output}} & 66.81 : 58.42 : 64.94 & 63.97 : 45.94 : 59.31\\
\textbf{\makecell{GECToR Output EC}} & 48.24 : 71.38 : 51.59 &50.50 : 61.09 : 52.31 \\ \hline
\textbf{\makecell{GEC-PD Output}} & 66.20 : 61.48 : 65.20 & 64.06 : 44.92 : 59.03\\
\textbf{\makecell{GEC-PD Output EC}} & 47.33 : 70.54 : 50.66 & 52.17 : 60.86 : 53.71\\ \hline
\textbf{\makecell{GRECO Model Output}} & 86.45 : 63.13 : 80.50 & 79.36 : 48.69 : 70.48\\ \hline
\end{tabular}
\caption{Quality Scores of the 2 GEC Tools output prediction, target Editor Corrections (EC) and the State-of-the-Art GEC Tool - GRECO \cite{qorib2023system} on the official BEA19 and CONLL14 competition.}
\label{tab:errant_scores_all}
\end{table*}

\begin{table*}[t]
\centering
\begin{tabular}{|l|c|c|}
\hline
\textbf{\makecell{Candidate Set}} & \makecell{\textbf{BEA19 Test}\\(P : R : $F_{0.5}$)} & \makecell{\textbf{CONLL14 Test}\\(P : R : $F_{0.5}$)} \\ \hline\hline
\textbf{\makecell{GECToR Output}} & 52.59 : 28.59 : 45.03 & 57.74 : 25.10 : 45.82\\
\textbf{\makecell{GECToR Output EC}} & 45.47 : 47.91 : 45.94 & 44.31 : 43.53 : 44.15 \\ \hline
\textbf{\makecell{GEC-PD Output}} & 49.88 : 26.37 : 42.33 & 56.49 : 23.13 : 43.85\\
\textbf{\makecell{GEC-PD Output EC}} & 45.90 : 48.31 : 46.36 & 46.14 : 42.64 : 45.39\\ \hline
%\textbf{\makecell{GRECO Model Output}} & 65.87 : 28.58 : 52.24 & 61.34 : 26.28 : 48.42\\ \hline
\end{tabular}
\caption{Quality Scores of the 2 GEC Tools output predictions and their final target Editor Corrections (EC) using the BEA19 and CONLL14 - Source Sentence EC as target reference.}
\label{tab:errant_scores_all_ourTRG}
\end{table*}

To overcome this issue, we calculate the quality scores for the 3 target correction sets and the GEC-Tool first-pass outputs on the official BEA19 and CONLL14 competitions and compare trends between the correction sets. We use the BEA19 competition website scorer\footnote{BEA19 GEC competition website - \url{https://codalab.lisn.upsaclay.fr/competitions/4057}} to evaluate the performance of BEA19 target corrections. Table \ref{tab:errant_scores_all} shows the quality scores for the GECToR and GEC-PD Tool output and the final editor target corrections (EC).

Similar trends are observed between the CONLL14 and BEA19 target correction sets. We observe a significant increase in Recall scores for the EC compared to the first-pass GEC Tool output. This indicates the final EC target contains additional post-edit corrections missed by the GEC Tool. The reduction in the precision score for EC is consistent with the 10 CONLL14 target corrections released by \citet{bryant2015far} since post-editing often leads to subjective paraphrasing and rewrite edits, which may not be present in the official competition target reference. The final EC obtained better Recall scores compared to the State-of-the-Art (SOA) GEC Tool - GRECO (as of writing this paper) \cite{qorib2023system} for both datasets. Observing similar quality score trends for the GEC Tool predictions and our target EC across both CONLL14 and BEA19 Test competition, and better Recall than the SOA GRECO tool, we can infer that the 3 target corrections collected by us in this work are of high quality.

We also use the GEC Inter Annotator Agreement (IAA) framework proposed by \citet{bryant2015far} to compare the target correction sets for both datasets with themselves to ensure better consistency and quality. The IAA framework states that the $F_{0.5}$ multi-reference score, used to evaluate a GEC Tool-vs-human corrections, can similarly evaluate human-vs-human corrections. When comparing multiple annotator corrections, a single correction set can be compared using the rest as references to get quality scores. The final IAA score is calculated as the average of all correction set scores. We use the ERRANT tool \cite{bryant2017automatic} to perform the IAA evaluation. We evaluate 3 target correction sets:

\begin{description}
\item[$A = \{A1-A10\}$] The 10 target corrections for CONLL14 by \citet{bryant2015far}.
\item[$C = \{c1,c2,c3\}$] The 3 CONLL14 target corrections collected by us.
\item[$B = \{b1,b2,b3\}$] The 3 BEA19 target corrections collected by us.
\end{description}

To compare IAA scores, we conduct a 1-vs-2 target correction set evaluation. For each correction in $A$, we randomly select 2 corrections from the remaining 9 as the reference. Scores for each correction in $B$ and $C$ are calculated using the remaining 2 corrections as target references. Table \ref{tab:IAA_Score} shows the average IAA scores for $A, B, C$ correction sets. We observe better Avg-IAA scores for the $C$ and $B$ correction sets collected by us in this work, compared to $A$.

\begin{table*}[t!]
\centering
\begin{tabular}{|l|c|c|}
\hline
\textbf{\makecell{Human Annotation Set}} & \makecell{\textbf{Reference Set and Size}} & \makecell{\textbf{IAA Score - }$F_{0.5}$} \\ \hline
$A1$ & $|\{RAND(2)\in \{A-A1\}| = 2$ & 36.21\\
$A2$ & $|\{RAND(2)\in \{A-A2\}| = 2$ & 45.48\\
$A3$ & $|\{RAND(2)\in \{A-A3\}| = 2$ & 46.72\\
$A4$ & $|\{RAND(2)\in \{A-A4\}| = 2$ & 40.54\\
$A5$ & $|\{RAND(2)\in \{A-A5\}| = 2$ & 46.01\\
$A6$ & $|\{RAND(2)\in \{A-A6\}| = 2$ & 50.85\\
$A7$ & $|\{RAND(2)\in \{A-A7\}| = 2$ & 42.72\\
$A8$ & $|\{RAND(2)\in \{A-A8\}| = 2$ & 49.46\\
$A9$ & $|\{RAND(2)\in \{A-A9\}| = 2$ & 52.0\\
$A10$ & $|\{RAND(2)\in \{A-A10\}| = 2$ & 48.57\\\hline
\textbf{Avg-IAA} $\{A\}$ & $\{A\}$, 2 & \textbf{45.85}\\\hline
$c1$ & $|\{C-c1\}| = 2$ & 54.11\\
$c2$ & $|\{C-c2\}| = 2$ & 57.36\\
$c3$ & $|\{C-c3\}| = 2$ & 59.14\\\hline
\textbf{Avg-IAA} $\{C\}$ & $\{C\}$, 2 & \textbf{56.87}\\\hline
$b1$ & $|\{B-b1\}| = 2$ & 57.94\\
$b2$ & $|\{B-b2\}| = 2$ & 59.39\\
$b3$ & $|\{B-b3\}| = 2$ & 59.81\\\hline
\textbf{Avg-IAA} $\{B\}$ & $\{B\}$, 2 & \textbf{59.05}\\\hline
\end{tabular}
\caption{Inter Annotator Agreement (IAA) scores for the different $A, B, C$ annotation sets using the ERRANT $F_{0.5}$ metric. RAND(n) represents a random selection of "n" items from the respective set.}
\label{tab:IAA_Score}
\end{table*}

%The Avg-IAA and official GEC competition quality scores indicate that the 3 target corrections that we collect for BEA19 and CONLL14 have similar or better quality than other public target corrections.
To ensure we choose strong GEC Tools (Section \ref{sec:gec}) to obtain first-pass output predictions, we compare the quality of the GEC Tool output and the subsequent human EC. We consider the Source Sentence EC (collected by us) as the target reference for the BEA19 and CONLL14 Test sets. The $F_{0.5}$ quality scores obtained in Table \ref{tab:errant_scores_all_ourTRG} show similar performance between the GECToR and GEC-PD Tool prediction output and the subsequent EC because of the variation in Precision and Recall scores. This indicates that GECToR and GEC-PD are strong first-pass GEC Tools.

\subsection{Impact of GEC Tools}\label{sec:impactgec}
Comparing the time-to-correct for the source sentence versus the GEC Tool output post-editing, we can quantify the impact of using GEC Tools in Text Editing.
\begin{table}[h!]
    \centering
    \begin{tabular}{|l|c|c|}
    \hline
        \textbf{\makecell{Sentence\\Source}} & \textbf{\makecell{Average Time\\per Sentence}} & \textbf{\makecell{Average Time\\per Word}} \\ \hline
        \makecell{Source\\Sentence} & 31.16 & 1.91 \\ \hline
        \makecell{GECToR\\Output} & 26.82 & 1.57 \\ \hline
        \makecell{GEC-PD\\Output} & 27.46 & 1.67 \\ \hline
    \end{tabular}
    \caption{The average time to correct (\textbf{in seconds}) for a sentence and word; correcting the source and after first-pass GEC Tool editing.}
    \label{tab:time_diff}
\end{table}

Quality scores presented in Table \ref{tab:errant_scores_all} show that the GEC Tool output EC has better values compared to the Source Sentence EC. In Table \ref{tab:time_diff}, we compare the time taken (in seconds) by a human editor to correct the source sentences with and without first-pass editing by a GEC tool. We observe that GEC Tools help in reducing the post-editing time by roughly $4$ seconds per sentence. Combined insights from these results indicate that incorporating GEC Tools in the text-editing workflow reduces editing time and generates better final target corrections. Thus, GEC Tools can help improve editor efficiency and overall productivity.

\section{Methodology}
We design statistical and neural network regression models for our post-editing effort in time (PEET) scorer. The scorer is trained to estimate the time-to-correct value for a source sentence given the target correction, using the number and type of edits and sentence property - Sentence Length, Correct/Incorrect.

The dataset that we collected contains 3 iterations for all 3 variations of the source - source (SRC), GEC Tool Model Output (MO) and post-edited target correction (TRG). Different training features in terms of edits and sentence structure can be selected and extracted from - SRC, MO and TRG triple (Appendix \ref{app:feat-select}).

Statistical PEET models performed as well as Neural models while allowing greater interpretability of training features (Appendix \ref{app:neural-peetmodel}). Also, models using features selected from $[MO, TRG]$ sentences performed better than models trained on fine-grained features from $[SRC, MO, TRG]$ sentences (Appendix \ref{app:SRC-MO-TRG-PEET}). Hence, we only discuss the features and results of the Statistical PEET Model trained using the $[MO, TRG]$ sentences here, referring to MO as the source.

\subsection{ERRANT Edit Feature Extraction}\label{sec:feat_errant}
\begin{figure}[h!]
    \centering
    \includegraphics[scale=0.43]{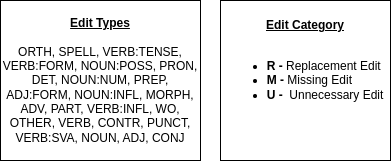}
    \caption{ERRANT edit category and types.}
    \label{fig:errant-eg}
\end{figure}

We use ERRANT \cite{bryant2017automatic} to align and extract edit features between the source and target corrections (Appendix \ref{app:errantGEC}). Apart from the edit category - Removal(R), Missing(M) and Unnecessary(U), the feature also includes the edit type. Figure \ref{fig:errant-eg} lists the different edit categories and their syntactic type generated by ERRANT.

We use the number and type of edits as features for our statistical models. Similar to the edit type hierarchy used by \citet{yuan2021multi}, considering category, type and their combination can provide 4, 25 or 55 edit features. For instance, if we only consider the 3 edit categories, then our 4 edit features are Replacement(R), Missing(M), Unnecessary(U) and Correct/Incorrect (binary feature). Using the 24 edit types (Figure \ref{fig:errant-eg}) and Correct/Incorrect gives us 25 edit features. Similarly, combining edit categories with their possible types, we get 55 edit features (see Table~\ref{tab:big-coeff} in Appendix~\ref{app:RC-others}). We train separate models for all three edit levels (4, 25, 55).

\subsection{PEET Scorer Models}\label{sec:models-loss}
We design Linear Regression (LR) and Support Vector Regression (SVR) models, for our PEET Scorer, using the ERRANT Edit count and different edit type levels (4, 25, 55), number of edited words, source and target sentence length as features. We also experimented with Neural Regression models, but they didn't perform better than statistical models (Results in Appendix \ref{app:neural-peetmodel}). We only discuss the results of the statistical PEET models here. The details of each model and the hyperparameters are presented in Appendix \ref{app:model-param}.

The PEET estimation task has a continuous range of prediction values - time (in seconds). We report the mean absolute error (MAE) and Pearson correlation ($r$) between the predicted time and the target time. We note that MAE does not take into account the sign of the error, while correlation does \cite{graham2015improving, tezcan2019estimating}, which is why we report correlation and use it to compare model performance.

\section{Experiment Results}
\subsection{Performance of the PEET Scorer}
The results for the Linear Regression (LR) and SVR PEET Scorer, with count of different edit feature levels (4,25,55), sentence word length and number of word edits as features (Section \ref{sec:feat_errant}), are presented in Table \ref{tab:TC-Result}.
\begin{table}[h!]
\resizebox{\columnwidth}{!}{%
\begin{tabular}{|l|l|l|l|}
\hline
\textbf{\begin{tabular}[c]{@{}l@{}}Statistical\\ Model\end{tabular}} & \textbf{\begin{tabular}[c]{@{}l@{}}Edit Feature\\Level\end{tabular}} & \textbf{$r$} & \textbf{MAE} \\ \hline
\multirow{3}{*}{\textbf{\begin{tabular}[c]{@{}l@{}}Linear\\ Regression\end{tabular}}} & 4 & 0.559 & 18.92 \\ \cline{2-4} 
 & 25 & \textbf{0.565} & 18.74  \\ \cline{2-4} 
 & 55 & 0.563 & 18.75 \\ \hline
\multirow{3}{*}{\textbf{SVR Linear}} & 4 & 0.558 & 16.40 \\ \cline{2-4} 
 & 25 & 0.564 & 16.19 \\ \cline{2-4} 
 & 55 & \textbf{0.565} & 16.15 \\ \hline
\end{tabular}
}
\caption{Average PEET estimation performance for the Statistical Models over 50 runs (different train-test data seed). The results are presented as the Pearson Correlation ($r$), Mean Absolute Error (MAE) loss.}
\label{tab:TC-Result}
\end{table}

The statistical models relying on edit type information (25,55 labels) performed better than using minimal substitution, deletion and insertion edit category labels (Figure \ref{fig:errant-eg}). This indicates that the type of edit has an impact on post-editing effort. We obtain a correlation of $r=0.565$ from the best models (LR-25 edit features).

\subsection{Impact of Error Types on Post-Edit Effort}\label{sec:peetfeat}

We follow the work by \citet{ye2021predicting}, using regression coefficients of a Linear Regression (LR) model to estimate the PEET impact of different edit features. To make the coefficients interpretable, we center and standardize all edit-features by subtracting the mean and dividing by the standard deviation (except the binary/categorical edit feature - Correct/Incorrect) \cite{schielzeth2010simple}.

\begin{table}[h!]
\resizebox{\columnwidth}{!}{%
\begin{tabular}{|c|c|c|c|c|c|}
\hline
\textbf{\begin{tabular}[c]{@{}c@{}}Model\\ Features\end{tabular}} & \textbf{\begin{tabular}[c]{@{}c@{}}Regression\\ Coefficient\end{tabular}} & \textbf{\begin{tabular}[c]{@{}c@{}}Model\\ Features\end{tabular}} & \textbf{\begin{tabular}[c]{@{}c@{}}Regression\\ Coefficient\end{tabular}} & \textbf{\begin{tabular}[c]{@{}c@{}}Model\\ Features\end{tabular}} & \textbf{\begin{tabular}[c]{@{}c@{}}Regression\\ Coefficient\end{tabular}} \\ \hline
\textbf{OTHER} & 10.15 & \textbf{ORTH} & 2.34 & \textbf{ADJ} & 0.97 \\ \hline
\textbf{PUNCT} & 4.55 & \textbf{CONJ} & 2.03 & \textbf{CONTR} & 0.78 \\ \hline
\textbf{PREP} & 4.03 & \textbf{MORPH} & 1.89 & \textbf{VERB:INFL} & 0.63 \\ \hline
\textbf{VERB} & 3.37 & \textbf{SPELL} & 1.87 & \textbf{PART} & 0.47 \\ \hline
\textbf{\begin{tabular}[c]{@{}c@{}}Sentence\\Correct\end{tabular}} & -3.31 & \textbf{ADV} & 1.79 & \textbf{ADJ:FORM} & 0.39 \\ \hline
\textbf{NOUN} & 3.23 & \textbf{VERB:FORM} & 1.66 & \textbf{NOUN:INFL} & -0.30 \\ \hline
\textbf{DET} & 3.08 & \textbf{WO} & 1.63 & \textbf{NOUN:POSS} & 0.25 \\ \hline
\textbf{NOUN:NUM} & 2.52 & \textbf{VERB:SVA} & 1.16 & - & - \\ \hline
\textbf{VERB:TENSE} & 2.35 & \textbf{PRON} & 1.10 & - & - \\ \hline
\end{tabular}%
}
\caption{The standardized regression coefficients of the LR model trained on the medium (25) edit features to measure the impact of each feature on PEET estimation.}
\label{tab:med-coeff}
\end{table}

The edit category \textit{OTHER}, which corresponds to paraphrasing or rewriting text, and modifying punctuation has the highest impact on post editing time. Deciding whether a particular sentence is incorrect also contributes significantly to the post-editing effort. The coefficients to study the impact of the 25 edit features are shown in Table \ref{tab:med-coeff}. Coefficients for the other edit granularities (4 and 55 labels) and all PEET sentence features are provided in Appendix \ref{app:RC-others}.

\subsection{PEET Scorer for GEC Quality Estimation}\label{sec:hjr-sets-ranking}

Since an efficient GEC Tool would reduce post-editing (PE) time, PE followed by PEET estimation can quantify the usability of a GEC Tool \cite{specia2011exploiting}. To study the correlation between cognitive, temporal and technical PE effort, we compare the PEET scorer rankings with human judgment rankings (HJR) (Section \ref{sec:PEE-GEC}) and Word Error Rate (Technical Effort) of GEC Tools. We evaluate the PEET-Linear Regression (25 Edit Features) Scorer (Section \ref{sec:feat_errant}) estimated ranking for 33 GEC Tools in 3 GEC HJR (Appendix \ref{app:HJR-PEET}).
\begin{itemize}
    \item \textit{Grundkiewicz-C14(EW)} - ranking of 12 GEC Tools that participated in the official CONLL-14 - GEC Task \cite{ng2014conll} by \citet{grundkiewicz2015human}.
    \item \textit{SEEDA-C14-All(TS)} - ranking of 15 newer and stronger GEC Tools on the CONLL-14 test dataset by \citet{kobayashi2024revisiting}. \textit{SEEDA-C14-NO(TS)} denotes the subset of 12 GEC tools without the 3 outliers.
    \item \textit{Napoles-FCE} and \textit{Napoles-Wiki} - ranking of 6 Seq2Seq GEC Tools on the FCE \cite{yannakoudakis2011new} and WikiEd \cite{grundkiewicz2014wiked} datasets by \citet{napoles2019enabling}.
\end{itemize}

\begin{table}[h!]
%\tiny
\resizebox{\columnwidth}{!}{%
\begin{tabular}{|l|ll|ll|}
\hline
\multirow{2}{*}{\textbf{Human Judgment Ranking}} & \multicolumn{2}{l|}{\textbf{PEET Metric}} & \multicolumn{2}{l|}{\textbf{WER}} \\ \cline{2-5} 
                                 & \multicolumn{1}{l|}{\textbf{$\rho$}}     & \textbf{$r$} & \multicolumn{1}{l|}{\textbf{$\rho$}}     & \textbf{$r$} \\ \hline
\textbf{Grundkiewicz - C14 (EW)} & \multicolumn{1}{l|}{0.48}           & 0.26         & \multicolumn{1}{l|}{0.28}           & 0.18         \\ \hline
\textbf{SEEDA - C14 - All (TS)}  & \multicolumn{1}{l|}{0.18}           & 0.63         & \multicolumn{1}{l|}{0.18}           & 0.65         \\ \hline
\textbf{SEEDA - C14 - NO (TS)}   & \multicolumn{1}{l|}{-0.1}           & -0.27        & \multicolumn{1}{l|}{-0.1}           & -0.33        \\ \hline
\textbf{Napoles - FCE}           & \multicolumn{1}{l|}{\textbf{-0.96}} & -0.94        & \multicolumn{1}{l|}{\textbf{-0.96}} & -0.88        \\ \hline
\textbf{Napoles - Wiki}          & \multicolumn{1}{l|}{\textbf{-0.71}} & -0.63        & \multicolumn{1}{l|}{\textbf{-0.93}} & -0.88        \\ \hline
\end{tabular}%
}
\caption{The correlation of our PEET model ranking with human-judgment rankings (HJR). We also provide the correlation of the HJR with the Word Edit Rate (WER) metric. Spearman ($\rho$) and Pearson ($r$) correlation scores are used for comparison. A high negative correlation indicates lower time-to-correct and WER score corresponding to a higher human judgment ranking.}
\label{tab:peet-human-wer}
\end{table}

The \textit{Grundkiewicz-C14} and \textit{SEEDA-C14} human ranking calculation was conducted using the Expected Wins (EW) \cite{Bojar2013FindingsOT} and TrueSkill (TS) \cite{herbrich2006trueskill} method, which tracks relative ranking based on a set-wise comparison of a subset of all GEC Tool corrections. The EW and TS rankings were selected for the final \textit{Grundkiewicz-C14} and \textit{SEEDA-C14} rankings, respectively. The \textit{Napoles - FCE} and \textit{Napoles - Wiki} human ranking addressed the issue of partial comparison and relative ranking for GEC Tools by using the partial ranking with scalars (PRWS) method \cite{sakaguchi2018efficient}, collecting a quality score (0-100) for each sentence to infer the final rankings.

Table \ref{tab:peet-human-wer} shows the Pearson ($r$) and Spearman ($\rho$) correlation scores of the HJRs with the PEET model ranking and the Word Error Rate (WER) \cite{snover2006study} (number of edits required to correct a GEC Tool prediction). The WER and PEET are calculated using untargeted references, which contributes to the lower alignment with perceived cognitive effort judgment.

We observe a good alignment (high negative correlation) between the PEET ranking and the \textit{Napoles} HJR and a poor alignment (positive correlation) with the other HJRs. The PEET ranking shows better alignment to HJRs that align with WER scores (Technical PE effort - Section \ref{sec:PEE-GEC}). We also observe that human quality rankings collected using PRWS align better with true human effort (WER) than those collected using TS or EW.

These results suggest that our PEET Scorer can estimate GEC Tool usability when output quality depends on further Post-Editing Effort (WER and type of edits) required to correct the tool output. Hence, performing PE to obtain the closest correction (lower WER) can improve GEC temporal effort estimation.

\section{Conclusion and Future Work}
Since we present the first study and dataset of Post-Editing Effort (PEET) in Time for GEC, our goal is to provide a baseline for future work in this area. Using our dataset, we quantified the editor efficiency and productivity when using GEC Tools for Text Editing. We extract various automated sentence properties and edit type features from the sentence correction pairs to train the PEET Scorer. Recently, there has been some work in the area of Grammar Error Explanation to define descriptive error types \cite{fei-etal-2023-enhancing, ye2025corrections} and use LLMs for error explanation \cite{song2023gee, li2025explanation}. As future work, the descriptive edits can be used as possible features for the PEET model. Finally, we observe that our PEET model works well for GEC Tool evaluation when the output quality is dependent on the Technical PE Effort (amount of edits). Studying actual cognitive effort for GEC post-editing and how it compares with technical and temporal effort is another interesting direction for future work.

\section*{Limitations}
One of the main limitations of Post-Editing (PE) Effort estimation is incorporating human annotation to evaluate GEC Tool performance, which can be expensive. However, PE allows us to quantify the true performance from a human-in-the-loop perspective. Currently, our work is limited to automated edit-type features generated by the ERRANT toolkit \cite{bryant2017automatic}.
Evaluating our PEET Scorer as a GEC quality estimation tool shows that it is effective when the correction quality is dependent on the technical post-editing effort. However, similar to work in Machine Translation, it is inconsistent with quality estimation based on perceived PE efforts.
Finally, we acknowledge that our work is limited to only the English language. Future work on post-editing GEC for other languages can show the impact of language type on PEET for GEC.

\section*{Acknowledgments}
We thank staff and colleagues at Scribendi Inc., Chatham, Ontario (\url{www.scribendi.com}) for the grant, input and feedback during the research and manuscript writing phases for this project. Resources used in this work were provided by the Province of Ontario, the Government of Canada through CIFAR, companies sponsoring the Vector Institute (\url{https://vectorinstitute.ai/partnerships/current-partners/}), the Natural Sciences and Engineering Council of Canada and a grant from IITP \& MSIT of Korea (No. RS-2024-00457882, AI Research Hub Project). 

\bibliography{custom}

\appendix

\section{Neural Regression Models for PEET Estimation}\label{app:neural-peetmodel}

\begin{table*}[h!]
    \centering
    \begin{tabular}{|c|c|}
    \hline
        \textbf{\makecell{Model Type}} & \textbf{\makecell{Input Format}} \\ \hline
        \makecell{Sentence Edit} & [MO] <mo-sentence> [TRG] <trg-
sentence>\\ \hline
        \makecell{Syntactic Variation} & <mo-constituency-parse> [TO] <trg-constituency-parse> \\ \hline
        \makecell{\#EW + Syntactic Variation} & \#EW - <mo-constituency-parse> [TO] <trg-constituency-parse> \\ \hline
        \makecell{\#EW + Syntax Structure} & \#EW - <trg-part-of-speech-tag>  \\ \hline
    \end{tabular}
    \caption{The training data format for the BERT and RoBERTa LM. The example considers a sentence pair - <mo-sentence> and <trg-
sentence> where "mo" is the Model Output correction made by a GEC Tool and the "trg" is the post-edited target correction for "mo". The special tokens [MO], [TRG] and [TO] denote sentence breaks in the input. \#EW denotes the number of edited words between mo and trg.}
    \label{tab:llm-input}
\end{table*}

Since semantics and syntax structure have been shown to impact PE effort \cite{tezcan2016detecting,bangalore2015role}, we also trained neural-LM PEET Scorer models using flattened constituency parse trees \cite{kitaev2018constituency} and part-of-speech syntax structure features for the source and target corrections, generated using the spaCy library \cite{spacy2}. 

\begin{table}[H]
%\small
\centering
\resizebox{\columnwidth}{!}{%
\begin{tabular}{|l|ll|ll|}
\hline
\multicolumn{1}{|c|}{\multirow{2}{*}{\textbf{Model Features}}} & \multicolumn{2}{c|}{\textbf{BERT-L}} & \multicolumn{2}{c|}{\textbf{RoBERTa-L}} \\ \cline{2-5} 
\multicolumn{1}{|c|}{} & \multicolumn{1}{l|}{$r$} & MAE & \multicolumn{1}{l|}{$r$} & MAE \\ \hline
\bfseries\makecell{Sentence Edit} & \multicolumn{1}{l|}{0.552} & 17.73 & \multicolumn{1}{l|}{\textbf{0.56}}  & 17.97 \\ \hline
\bfseries\makecell{Syntactic\\Variation} & \multicolumn{1}{l|}{0.528} & 19.35 & \multicolumn{1}{l|}{\textbf{0.564}} & 18.05 \\ \hline
\bfseries\makecell{\#EW + Syntactic\\Variation} & \multicolumn{1}{l|}{\textbf{0.564}} & 17.16 & \multicolumn{1}{l|}{0.561} & 16.88 \\ \hline
\bfseries\makecell{\#EW + Syntax\\Structure} & \multicolumn{1}{l|}{\textbf{0.565}} & 18.57 & \multicolumn{1}{l|}{0.565} & 18.74 \\ \hline
\end{tabular}
}
\caption{Performance of Neural PEET models using different sequence model features over 5 runs. The results are shown as Pearson Correlation ($r$) and Mean Absolute Error (MAE) loss.} \label{tab:transfResults}
\end{table}

Pretrained LMs can also capture syntax structure internally \cite{dai2021does}, so we also train neural-LM models using only source-target sentence embeddings as features to estimate PEET. Since the statistical models work as well as Neural models, while being faster and more interpretable, we consider them for the PEET Scorer in the main paper. We describe the features (Table \ref{tab:llm-input}) and results of the Neural PEET (Table \ref{tab:transfResults}) model here.

\section{GEC Evaluation File Example and Format}\label{app:errantGEC}

\begin{table}[!h]
\begin{tabular}{p{1.5cm} p{5cm}}
\centering
\textbf{Source :}& Surrounded by such concerns, it is very likely that we \underline{are} distracted to worry about these problems.\\
\textbf{Target :}& Surrounded by such concerns, it is very likely that we \underline{will be too} distracted to worry about these problems. \\
\textbf{Model\newline Output :} & Surrounded by such concerns, it is very likely that we are distracted \underline{from worrying} about these problems.
\end{tabular}
\caption{\textit{Source}, \textit{Target} and example \textit{Model Output} made by a GEC Tool.}
\label{tab:gec-example}
\end{table}

The evaluation of a GEC Tool requires a Source (S), Target (T) and Model Output (MO) sentence. Table \ref{tab:gec-example} gives an example of such a triple. GEC evaluation generates M2 file for a pair of sentences (e.g., S and T), which lists the edits that can convert sentence S into sentence T and the positions of those edits. The evaluation process generates two M2 files : (Source - Target) and (Source - Model Output). The M2 edits are compared to evaluate the Model Output quality.\\

\small
$\bullet$ \textit{Source-Target} M2 File:\\
\texttt{S Surrounded by such concerns , it is very likely that we are distracted to worry about these problems .\\
A 13 14|||R:OTHER|||and|||REQUIRED||| -NONE-|||0\\
A 11 12|||R:VERB:TENSE|||will be|||REQUIRED||| -NONE-|||1\\
A 12 12|||M:ADV|||too|||REQUIRED||| -NONE-|||1}
\\$\bullet$ \textit{Source-Model Output} M2 File:\\
\texttt{S Surrounded by such concerns , it is very likely that we are distracted to worry about these problems .\\
A 13 14|||R:PART|||from|||REQUIRED||| -NONE-|||0\\
A 14 15|||R:VERB:FORM|||worrying||| REQUIRED||| -NONE-|||0\\
}

\normalsize
The M2 file format was part of the GEC-M2 Scorer evaluation tool proposed by \citet{dahlmeier2012better}. The tool generates an alignment and detects atomic edits between a pair of sentences. Further improvement to the M2 tool was done by \citet{bryant2017automatic}, resulting in the ERRANT evaluation tool. The ERRANT tool retained the overall M2 file format, utilizing syntactic and linguistic features to extract better-aligned and tagged edits between 2 sentences (as shown above).

\section{Predictive Model Parameters}\label{app:model-param}
We train different statistical and neural predictive models to estimate the post-editing temporal effort. We use this section to describe the predictive models as well as the training parameters for the regression task.
\begin{description}
    \item[Linear Regression: ]We use the Linear Regression (LR) model provided by the Scikit-Learn library\footnote{\url{https://scikit-learn.org/stable/modules/generated/sklearn.linear_model.Ridge.html}}. To keep the weights of the features from getting arbitrarily high, we used the RidgeLinear model that also adds an L2 Regularizer to the model. We trained the model with default training parameters and $alpha=1.0$.
    \item[Support Vector Regression: ]We also train Support Vector Regression (SVR) models from scikit-learn with the default training parameters and the "linear" kernel.
    \item[BERT, RoBERTa Neural Models:] To train neural predictive models, we fine-tuned the BERT-Large \cite{devlin2018bert} and RoBERTa-Large \cite{liu2019roberta} with a regression head. The models were trained using the Pfeiffer bottleneck adapters \cite{pfeiffer2020adapterfusion} which allowed us to reduce the training time. We utilized the AdapterHub library\footnote{\url{https://adapterhub.ml/}} for training the models with the default Pfeiffer adapter configuration \cite{pfeiffer2020adapterhub}. Training was done for 50 epochs with a 10-epoch and $.05$ loss threshold early stopping. A learning rate of $1e-04$ was used. To train the models for the regression task, we added a one-label regression head and used the mean-square-error loss (MSELoss), which is part of the Huggingface\footnote{\url{https://huggingface.co/}} training pipeline.
\end{description}

\section{Different Sources for Training Feature Selection and Extraction}\label{app:feat-select}

\begin{figure}[h!]
    \centering
    \includegraphics[scale=0.33]{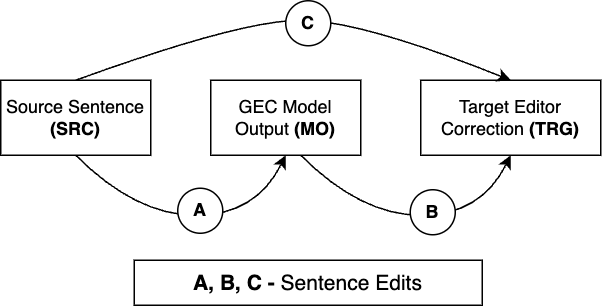}
    \caption{Sentence correction edits extracted using the ERRANT toolkit.}
    \label{fig:edit_feat}
\end{figure}

Our dataset has 3 iterations for each source sentence. We have the original sentence - source (SRC), the first-pass correction by a GEC Tool - Model Output (MO) and the final targeted editor correction - target (TRG).
%Out of our 3 correction sets, 2 sets have a first-pass correction performed by a GEC Tool, and 1 set has the correction for the source sentence.For source sentence correction, we consider the Model Output to be the Source Sentence ($MO = SRC$). This can be summarized as using a GEC Tool that makes $0$ edits to a source.
Figure \ref{fig:edit_feat} shows the 3 iterations for the source sentence. Each arc represents a sentence transition pairing and can be used to extract intermediate edit features. To extract features, the following sentence pairings can be considered: [MO], [SRC - MO], [\textbf{MO - TRG}], [\textbf{SRC - MO - TRG}].
Post-editing features, from different levels, can be extracted from the $SRC-MO-TRG$ and $MO-TRG$ sentence pairings. Considering the source sentence as a feature can further separate target edits into ignored and incorrect edits.
\begin{itemize}
\item \textbf{SRC - MO - TRG}: We consider and extract the set of edits - A and C (Figure \ref{fig:edit_feat}) for the model features. We further use these edits to create 2 categories - Incorrect and Ignored edits.
\begin{itemize}
    \item Incorrect: $|A - C|$
    \item Ignored: $|C - A|$
\end{itemize}

\item \textbf{MO - TRG}: We consider only edit set - B (Figure \ref{fig:edit_feat}) as the input for the trained models.
\end{itemize}

We found that the performance of models trained on these 2 feature sources was comparable (Appendix \ref{app:SRC-MO-TRG-PEET}). This also indicates that the PEET Scorer can estimate time-to-correct from the post-editing correction stage - B. We only present and discuss the results of the model trained using the $MO-TRG$ sentence features in the main paper. Results for the $[SRC-MO-TRG]$ Scorer are presented in Appendix \ref{app:SRC-MO-TRG-PEET}.

\section{PEET Scorer using SRC, MO and TRG Sentence Features}\label{app:SRC-MO-TRG-PEET}

\begin{table}[H]
\centering
\begin{tabular}{|l|ll|ll|}
\hline
\multicolumn{1}{|c|}{\multirow{2}{*}{\textbf{\shortstack{Model\\ Features}}}} & \multicolumn{2}{c|}{\textbf{BERT-L}} & \multicolumn{2}{c|}{\textbf{RoBERTa-L}} \\ \cline{2-5} 
\multicolumn{1}{|c|}{} & \multicolumn{1}{l|}{$r$} & MAE & \multicolumn{1}{l|}{$r$} & MAE \\ \hline
\bfseries\makecell{\shortstack{Sentence\\Edit}} & \multicolumn{1}{l|}{0.513} & 19.10 & \multicolumn{1}{l|}{0.54} & 17.82 \\ \hline
\end{tabular}%
\caption{Neural PEET model performance over 5 runs using the source (SRC), GEC Tool Model Output (MO) and Target Correction (TRG) sentence features. The results are shown as Pearson Correlation ($r$) and Mean Absolute Error (MAE) loss.}
\label{tab:neural-peet-smotrg}
\end{table}

\begin{table}[H]
\centering
\begin{tabular}{|l|l|l|l|}
\hline
\textbf{\begin{tabular}[c]{@{}l@{}}Statistical\\ Model\end{tabular}} & \textbf{\begin{tabular}[c]{@{}l@{}}Edit Feature\\Level\end{tabular}} & \textbf{$r$} & \textbf{MAE} \\ \hline
\multirow{2}{*}{\textbf{\begin{tabular}[c]{@{}l@{}}Linear\\ Regression\end{tabular}}} & 10 & 0.558 & 18.92 \\ \cline{2-4} 
%& 106 & \textbf{0.565} & 18.74 \\ \cline{2-4} 
 & 106 & 0.557 & 18.89 \\ \hline
\multirow{3}{*}{\textbf{SVR Linear}} & 10 & 0.556 & 16.39 \\ \cline{2-4} 
%& 25 & 0.564 & 16.19 \\ \cline{2-4}
 & 106 & 0.561 & 16.21 \\ \hline
\end{tabular}%
\caption{PEET Statistical Model performance over 50 runs (different train-test data seed) using Incorrect and Ignored separated Edit features (Appendix \ref{app:feat-select}) extracted from SRC, MO and TRG sentence triples. The results are presented as the Pearson Correlation ($r$), Mean Absolute Error (MAE) loss.}
\label{tab:PEET-SRCMOTRG}
\end{table}

\section{GEC Post Editing Instructions and Survey Example}\label{app:survey}

\begin{figure}[H]
\includegraphics[width=\linewidth]{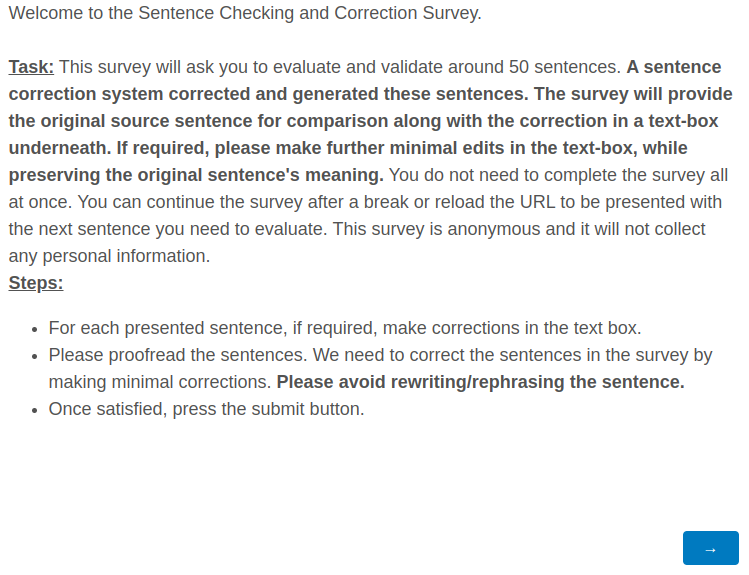}
\caption{Survey instructions for the editor to perform post editing, and obtain target corrections for our dataset.}
\label{fig:surv_inst}
\end{figure}

\begin{figure}[H]
\includegraphics[width=\linewidth]{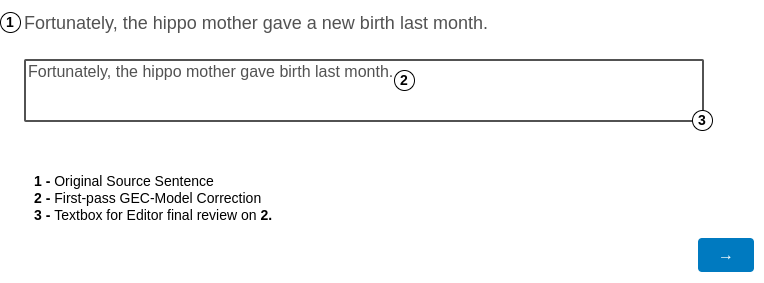}
  \caption{Example source sentence and its first-pass edit from the Survey. The editor can make further improvements in the text box. Submitting the final target correction.}
  \label{fig:survey_example}
\end{figure}

\section{Feature Impact on Post-Editing Time using Regression Coefficients}\label{app:RC-others}

\begin{table*}[t!]
\resizebox{\textwidth}{!}{%
\begin{tabular}{|c|c|c|c|c|c|c|c|c|c|}
\hline
\textbf{\begin{tabular}[c]{@{}c@{}}Model\\ Features\end{tabular}} & \textbf{\begin{tabular}[c]{@{}c@{}}Regression\\ Coefficient\end{tabular}} & \textbf{\begin{tabular}[c]{@{}c@{}}Model\\ Features\end{tabular}} & \textbf{\begin{tabular}[c]{@{}c@{}}Regression\\ Coefficient\end{tabular}} & \textbf{\begin{tabular}[c]{@{}c@{}}Model\\ Features\end{tabular}} & \textbf{\begin{tabular}[c]{@{}c@{}}Regression\\ Coefficient\end{tabular}} & \textbf{\begin{tabular}[c]{@{}c@{}}Model\\ Features\end{tabular}} & \textbf{\begin{tabular}[c]{@{}c@{}}Regression\\ Coefficient\end{tabular}} & \textbf{\begin{tabular}[c]{@{}c@{}}Model\\ Features\end{tabular}} & \textbf{\begin{tabular}[c]{@{}c@{}}Regression\\ Coefficient\end{tabular}} \\ \hline
\textbf{R:OTHER} & 7.73 & \textbf{M:DET} & 2.03 & \textbf{M:VERB} & 1.49 & \textbf{U:VERB} & 1.07 & \textbf{M:ADJ} & 0.36 \\ \hline
\textbf{U:OTHER} & 4.53 & \textbf{M:OTHER} & 1.98 & \textbf{R:VERB:FORM} & 1.48 & \textbf{M:ADV} & 0.93 & \textbf{R:CONJ} & 0.30 \\ \hline
\textbf{\begin{tabular}[c]{@{}c@{}}Sentence Correct\end{tabular}} & -3.11 & \textbf{R:DET} & 1.94 & \textbf{U:PUNCT} & 1.36 & \textbf{U:ADJ} & 0.79 & \textbf{U:NOUN:POSS} & 0.29 \\ \hline
\textbf{R:PREP} & 2.85 & \textbf{M:PREP} & 1.93 & \textbf{U:ADV} & 1.32 & \textbf{R:VERB:INFL} & 0.58 & \textbf{U:VERB:TENSE} & 0.25 \\ \hline
\textbf{R:PUNCT} & 2.84 & \textbf{R:MORPH} & 1.77 & \textbf{M:VERB:TENSE} & 1.32 & \textbf{R:ADV} & 0.53 & \textbf{M:PART} & 0.18 \\ \hline
\textbf{M:PUNCT} & 2.80 & \textbf{U:PREP} & 1.69 & \textbf{M:VERB:FORM} & 1.29 & \textbf{M:NOUN:POSS} & 0.52 & \textbf{U:PART} & 0.10 \\ \hline
\textbf{R:VERB} & 2.71 & \textbf{R:SPELL} & 1.66 & \textbf{U:NOUN} & 1.26 & \textbf{R:ADJ} & 0.51 & \textbf{R:NOUN:POSS} & -0.06 \\ \hline
\textbf{R:NOUN} & 2.64 & \textbf{U:CONJ} & 1.64 & \textbf{M:NOUN} & 1.22 & \textbf{M:PRON} & 0.49 & \textbf{U:PRON} & 0.06 \\ \hline
\textbf{R:NOUN:NUM} & 2.32 & \textbf{U:DET} & 1.62 & \textbf{R:PRON} & 1.14 & \textbf{R:PART} & 0.42 & \textbf{U:VERB:FORM} & 0.05 \\ \hline
\textbf{R:ORTH} & 2.22 & \textbf{R:WO} & 1.58 & \textbf{R:VERB:SVA} & 1.11 & \textbf{R:ADJ:FORM} & 0.41 & \textbf{M:CONTR} & 0.02 \\ \hline
\textbf{R:VERB:TENSE} & 2.08 & \textbf{M:CONJ} & 1.52 & \textbf{U:CONTR} & 1.10 & \textbf{R:NOUN:INFL} & -0.37 & \textbf{R:CONTR} & 0.02 \\ \hline
\end{tabular}%
}
\caption{The standardized regression coefficients of the LR model trained on all the big (55) edit features to measure the impact of each feature on PEET estimation.}
\label{tab:big-coeff}
\end{table*}

We utilize the regression coefficients of a Ridge-Linear Regression model to quantitatively calculate the impact of different edit type features on the time-to-correct value (Section \ref{sec:peetfeat}). We provide the estimated impact of all edit types here.

\begin{table}[H]
\centering
\begin{tabular}{|l|l|}
\hline
\textbf{Model Features} & \textbf{\begin{tabular}[c]{@{}l@{}}Regression \\ Coefficient\end{tabular}} \\ \hline
\textbf{Substitutions (R)} & 14.05 \\ \hline
\textbf{Deletions (U)} & 6.71 \\ \hline
\textbf{Insertions (M)} & 5.28 \\ \hline
\textbf{Sentence Correct (C)} & -2.33 \\ \hline
\end{tabular}
\caption{The standardized regression coefficients of the LR model trained on the small (4) edit features to measure the impact of each feature on PEET estimation.}
\label{tab:small-coeff}
\end{table}

%Table \ref{tab:big-coeff} lists the coefficients for the 55 edit features, and we notice the greatest impact of determining whether the sentence is incorrect and making rephrasing/rewriting edits. Table \ref{tab:small-coeff} lists the regression coefficients for the 4 edit types, and we notice that word substitution edits had the highest impact on temporal Post Editing effort.

\begin{table}[H]
\resizebox{\linewidth}{!}{%
\centering
\begin{tabular}{|l|l|l|}
\hline
\textbf{Model Features} & \textbf{\begin{tabular}[c]{@{}l@{}}PEET\\ Correlation\end{tabular}}  & \textbf{\begin{tabular}[c]{@{}l@{}}Regression\\ Coefficient\end{tabular}} \\ \hline
\textbf{\# of words in TRG} & 0.43 & 14.07 \\ \hline
\textbf{Substitutions (R)} & 0.47 & 6.76 \\ \hline
\textbf{\# of Edited Words} & 0.52 & 6.46 \\ \hline
\textbf{\# of Words in MO} & 0.43 & -5.86 \\ \hline
\textbf{Deletions (U)} & 0.32 & 3.85 \\ \hline
\textbf{\begin{tabular}[c]{@{}l@{}}Sentence\\Correct (C)\end{tabular}} & -0.3 & -2.63 \\ \hline
\textbf{Insertions (M)} & 0.28 & 0.66 \\ \hline
\end{tabular}
}%
\caption{The correlation of the features used to train the small-edits(4) Linear Regression (LR) model in Table \ref{tab:TC-Result}. We also list the standardized regression coefficients to measure the impact of each feature on PEET estimation.}
\label{tab:allfeat-coeff}
\end{table}

%Finally, we show the coefficient weights and the correlation of sentence property features with the time-to-correct value in Table \ref{tab:allfeat-coeff}.

%TESTING ENDS HERE.

\section{PEET Scorer Ranking and Comparison of GEC Tools with Human Judgment Rankings}\label{app:HJR-PEET}

We evaluate and rank 33 different GEC Tools and correction sets, part of 3 GEC Human Judgment Rankings, to estimate the quality of our PEET Scorer (Section \ref{sec:hjr-sets-ranking}). We list all the GEC Tools along with the Human Judgment and PEET Scorer rankings here.

\begin{table}[H]
    \centering
    \begin{tabular}{|l|c|c|c|}
    \hline
        \textbf{\makecell{Model\\Name}} & \textbf{\makecell{HJR\\Score}} & \textbf{\makecell{PEET\\Score}} & \textbf{\makecell{PEET\\Ranking}} \\ \hline
marian & 76.99 & 21.82 & 1\\\hline
lstm-r & 74.48 & 22.45 & 3\\\hline
lstm & 74.3 & 22.39 & 2\\\hline
nus & 73.94 & 22.47 & 4\\\hline
transformer & 73.9 & 22.79 & 5\\\hline
amu & 70.68 & 23.27 & 6\\\hline
input & 68.15 & 23.3 & 7\\\hline
    \end{tabular}
    \caption{PEET Scorer estimated average time-to-correct per sentence and ranking for 7 GEC Tool corrections on the FCE dataset (1936 Sentences), along with their Human Judgment Ranking (HJR), presented in \textit{Napoles-FCE} \cite{napoles2019enabling} (Section \ref{sec:hjr-sets-ranking}). The 7 GEC Tools consist of Seq2Seq Neural Models.}
    \label{tab:hjr-peet-napole-fce}
\end{table}

\begin{table}[H]
    \centering
    \begin{tabular}{|l|c|c|c|}
    \hline
        \textbf{\makecell{Model\\Name}} & \textbf{\makecell{HJR\\Score}} & \textbf{\makecell{PEET\\Score}} & \textbf{\makecell{PEET\\Ranking}} \\ \hline
lstm-r & 78.27 & 27.61 & 2\\\hline
lstm & 77.73 & 27.61 & 1\\\hline
amu & 75.98 & 28.35 & 5\\\hline
input & 75.89 & 27.72 & 3\\\hline
marian & 75.8 & 30.52 & 7\\\hline
nus & 75.78 & 28.34 & 4\\\hline
transformer & 71.53 & 29.77 & 6\\\hline
    \end{tabular}
    \caption{PEET Scorer estimated average time-to-correct per sentence and ranking for 7 GEC Tool corrections on the WikiEd dataset (1984 Sentences), along with their Human Judgment Ranking (HJR), presented in \textit{Napoles-Wiki} \cite{napoles2019enabling} (Section \ref{sec:hjr-sets-ranking}). The 7 GEC Tools consist of Seq2Seq Neural Models.}
    \label{tab:hjr-peet-napole-wiki}
\end{table}

Table \ref{tab:hjr-peet-napole-fce}-\ref{tab:hjr-peet-napole-wiki} list the estimation scores for the 6 Seq2Seq GEC Tools ranked by \citet{napoles2019enabling}. The chosen models were AMU \cite{junczys2016phrase}, LSTM/LSTM-R \cite{klein2018opennmt}, Marian \cite{sennrich2017university}, NUS \cite{chollampatt2018multilayer}, and, Transformer \cite{vaswani2017attention}. 

\begin{table}[h!]
    \centering
    \begin{tabular}{|l|c|c|c|}
    \hline
        \textbf{\makecell{Model\\Name}} & \textbf{\makecell{HJR\\Score}} & \textbf{\makecell{PEET\\Score}} & \textbf{\makecell{PEET\\Ranking}} \\ \hline
AMU & 0.628 & 25.8 & 8\\\hline
RAC & 0.566 & 26.61 & 13\\\hline
CAMB & 0.561 & 26.34 & 11\\\hline
CUUI & 0.55 & 25.91 & 9\\\hline
POST & 0.539 & 26.28 & 10\\\hline
UFC & 0.513 & 24.56 & 2\\\hline
PKU & 0.506 & 25.63 & 6\\\hline
UMC & 0.495 & 25.72 & 7\\\hline
IITB & 0.485 & 24.67 & 3\\\hline
SJTU & 0.463 & 24.84 & 4\\\hline
INPUT & 0.456 & 24.53 & 1\\\hline
NTHU & 0.437 & 26.6 & 12\\\hline
IPN & 0.3 & 25.62 & 5\\\hline
    \end{tabular}
\caption{PEET Scorer estimated average time-to-correct per sentence and ranking for 12 GEC Tool corrections on the CONLL14 dataset (1312 Sentences), along with their Human Judgment Ranking (HJR), presented in \textit{Grundkiewicz-C14(EW)} \cite{grundkiewicz2015human} (Section \ref{sec:hjr-sets-ranking}). The 12 GEC Tools consist primarily of rule-based and statistical machine translation architecture.}
\label{tab:hjr-peet-napole-grund}
\end{table}

Table \ref{tab:hjr-peet-napole-grund} lists the quality judgment for the 12 GEC Tools that participated in the CONLL14 GEC Task \cite{ng2014conll} performed by \citet{grundkiewicz2015human}. AMU \cite{junczys2014amu}, CAMB \cite{felice2014grammatical}, CUUI \cite{rozovskaya2014illinois}, IITB \cite{kunchukuttan2014tuning}, IPN \cite{hernandez2014conll}, NARA \cite{ng2014conll}, NTHU \cite{wu2014nthu}, PKU \cite{zhang2014unified}, POST \cite{lee2014postech}, RAC \cite{borocs2014racai}, SJTU \cite{wang2014grammatical}, UFC \cite{gupta2014grammatical}, and UMC \cite{wang2014factored}.

\begin{table}[h!]
    \begin{tabular}{|c|c|c|c|}
    \hline
        \textbf{\makecell{Model\\Name}} & \textbf{\makecell{HJR\\
        Score}} & \textbf{\makecell{PEET\\Score}} & \textbf{\makecell{PEET\\Ranking}} \\ \hline
REF-F & 0.992 & 30.53 & 15\\\hline
GPT-3.5 & 0.743 & 26.04 & 14\\\hline
T5 & 0.179 & 24.37 & 10\\\hline
TransGEC & 0.175 & 23.54 & 3\\\hline
REF-M & 0.067 & 24.04 & 8\\\hline
BERT-Fuse & 0.023 & 23.61 & 4\\\hline
\makecell{Riken-\\Tohoku} & -0.001 & 23.36 & 2\\\hline
PIE & -0.034 & 23.66 & 6\\\hline
LM-Critic & -0.163 & 24.37 & 9\\\hline
\makecell{Template\\GEC} & -0.168 & 25.21 & 13\\\hline
\makecell{GECToR-\\BERT} & -0.178 & 23.78 & 7\\\hline
UEDIN-MS & -0.179 & 23.36 & 1\\\hline
\makecell{GECToR-\\Ens}& -0.234 & 23.62 & 5\\\hline
BART & -0.3 & 24.75 & 12\\\hline
INPUT & -0.992 & 24.53 & 11\\\hline
    \end{tabular}
    \caption{PEET Scorer estimated average time-to-correct per sentence and ranking for 15 GEC Tool corrections on the CONLL14 dataset (1312 Sentences), along with their Human Judgment Ranking (HJR), presented in \textit{SEEDA-C14-All(TS)} \cite{kobayashi2024revisiting} (Section \ref{sec:hjr-sets-ranking}). The 15 GEC Tools consist of strong SOA Neural Models.}
    \label{tab:hjr-peet-seeda}
\end{table}

Table \ref{tab:hjr-peet-seeda} lists the recent GEC Tools evaluated by \citet{kobayashi2024revisiting}. GPT-3.5 \cite{coyne2023analyzing}, T5 \cite{rothe2021simple}, TransGEC \cite{fang2023transgec}, BERT-Fuse \cite{kaneko2020encoder}, Riken-Tohoku \cite{kiyono2019empirical}, PIE \cite{awasthi2019parallel}, LM-Critic \cite{yasunaga2021lm}, TemplateGEC \cite{li2023templategec}, GECToR-BERT \cite{omelianchuk2020gector}, UEDIN-MS \cite{grundkiewicz2019neural}, GECToR-Ens \cite{tarnavskyi2022ensembling}, BART \cite{lewis2020bart}.

\end{document}